# Ask to Know More: Generating Counterfactual Explanations for Fake Claims

Shih-Chieh Dai[†*]
sjdai@utexas.edu
The University of Texas at Austin
Austin, TX, USA

Yi-Li Hsu[†]
yili.hsu@iis.sinica.edu.tw
Institute of Information Science, Academia Sinica
Taipei, Taiwan
Department of Computer Science, National Tsing Hua University
Hsin Chu, Taiwan

Aiping Xiong
axx29@psu.edu
The Pennsylvania State University
University Park, PA, USA

Lun-Wei Ku
lwku@iis.sinica.edu.tw
Institute of Information, Academia Sinica
Taipei, Taiwan

## ABSTRACT

Automated fact-checking systems have been proposed that quickly provide veracity prediction at scale to mitigate the negative influence of fake news on people and on public opinion. However, most studies focus on veracity classifiers of those systems, which merely predict the truthfulness of news articles. We posit that effective fact checking also relies on people's understanding of the predictions. We propose elucidating fact-checking predictions using counterfactual explanations to help people understand why a specific piece of news was identified as fake.

In this work, generating counterfactual explanations for fake news involves three steps: asking good questions, finding contradictions, and reasoning appropriately. We frame this research question as contradicted entailment reasoning through question answering (QA). We first ask questions towards the false claim and retrieve potential answers from the relevant evidence documents. Then, we identify the most contradictory answer to the false claim by use of an entailment classifier. Finally, a counterfactual explanation is created using a matched QA pair with three different counterfactual explanation forms. Experiments are conducted on the FEVER dataset for both system and human evaluations. Results suggest that the proposed approach generates the most helpful explanations compared to state-of-the-art methods. Our code and data is publicly available. [1]

## 1 INTRODUCTION

Since the 2016 U.S. presidential election campaign, there has been an increasing awareness of the critical impact of fake news on our life. Misinformation also influences us dramatically, not only politically but also in terms of health, especially in the global COVID-19 pandemic. To combat misinformation, government and private institutes have put substantial effort into debunking misinformation and mitigating its influence via, for instance, fact-checking websites such as PolitiFact, Snopes, and FactCheck. These websites hire experts to fact-check the information, and also provide comprehensive reports for the fact-checked results. Although experts do provide high quality and precise results, such fact-checking processes are labor-intensive and time-consuming, which are not effective in combating the rapid spread of misinformation.

Recently, counterfactual explanations have received attention in the XAI research community. A counterfactual expression is defined as the result of doing something that is counter to facts [9]. In other words, there is a contradictory relationship between fact and counter-fact. For example, suppose someone just lost a footrace: he may conjecture thus: "If I had slept better last night, I would have won first place." Counterfactuals in XAI have been widely used for providing explanations for AI models [20]. One example of a counterfactual explanation can refer to Figure 1. Moreover, AI explanations have been shown to improve participants' perceived accuracy ratings of fake news [18]. Thus, we argue that generating a human-interpretable explanation is essential.

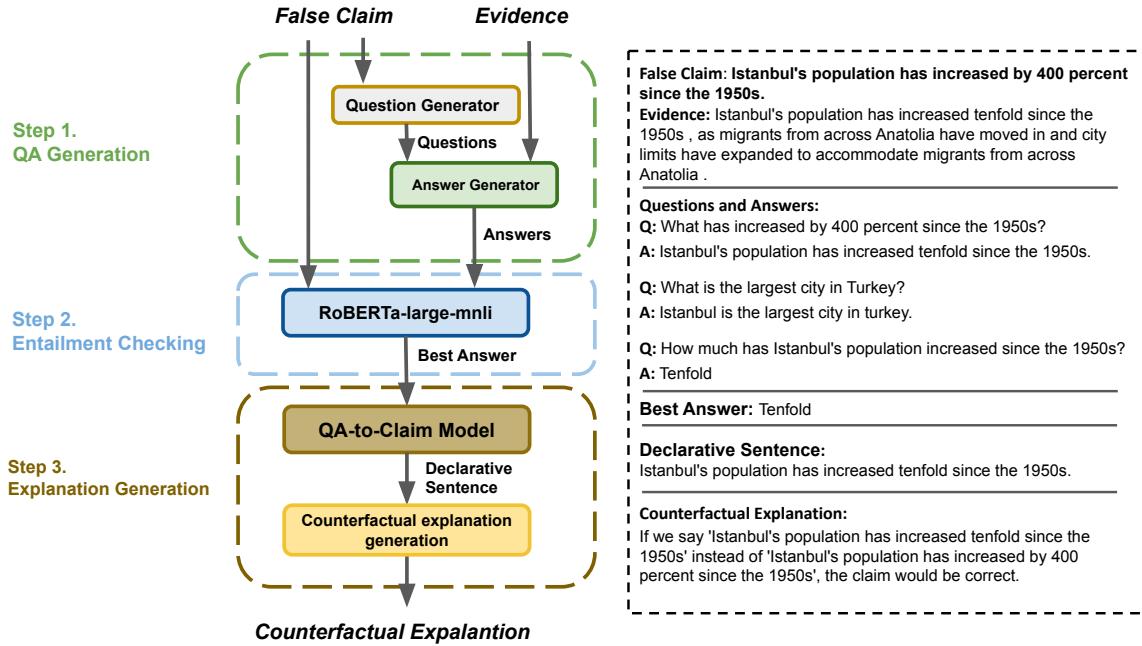

Figure 1: Overview of counterfactual explanation generation framework

However, studies have put little focus on whether counterfactual XAI is feasible for fake news explanations. In addition, it has been argued that not all counterfactuals are equally helpful for human comprehension. We seek to determine how counterfactuals influence the effectiveness of fake news explanations. Hence we study the following four research questions:

- RQ1: How can we generate a good counterfactual explanation for a given fake claim?
- RQ2: Do different types of counterfactual explanations (i.e., affirmative, negative, and mixed) vary in best explaining why a piece of news is fake?
- RQ3: How do counterfactual explanations best explain why a piece of news is fake compared to other state-of-the-art explanations?
- RQ4: Does an individual's familiarity (familiar vs. unfamiliar) with misinformation impact the effectiveness of counterfactual explanations?

Then, we formulate the following hypotheses when conducting the experiments:

(1) True statements that best refute the mistake in the fake claim tend to contain the most contradictory semantic meanings compared to the fake claim.
(2) Counterfactual explanations of fake news report higher overall acceptability compared to summarization-based explanations.

In this paper, we study the extraction of counterfactual information and explore the acceptability of counterfactual explanations of fake news compared to other summarization baselines.

We can divide fake news research into two main directions: identification and mitigation [37]. The goal of identification is to detect fake news. Mitigation-related research is to explore the solutions for mitigating the influence of fake news. Our scope is to generate the explanation for detected fake news articles, and thus we belong to mitigation-related research. Overall, we make the following contributions:

(1) We integrate the advantage of a question-answering model and a textual entailment model, propose a novel method to generate counterfactual information with 70% correctness, and show its usability under such performance.
(2) We propose three different counterfactual explanation forms and conduct human evaluations to compare their acceptability on the FEVER dataset.
(3) We show experimental results which strongly support that automatically-generated counterfactual explanations of fake news are more acceptable than summarization-based explanations.
(4) We show that counterfactual explanations are robust to system errors.

## 2 BACKGROUND
### 2.1 XAI
Explainable artificial intelligence (XAI) has drawn attention from academic researchers in recent years. To break the stereotype of AI

as a black box technology, researchers have proposed various methods to provide explainability for machine learning models. Common methodologies in natural language processing (NLP) for providing explanations are visualizing the attention score of the words [40] and using HeatMap [35]. These approaches benefit both researchers and stakeholders. For example, the researcher can improve the model performance with the assistance of explanations. Second, these approaches explain the rationale of the machine learning systems so that stakeholders can interpret the results of the machine learning model. Moreover, XAI also contributes to less bias in machine learning models [25] since it enhances the transparency of ML decision making and reveals bias in dataset [1]. However, for non-practitioners, such explanations are often counterintuitive. It is thus essential to generate human-interpretable explanations.

With the growing dependence on machine learning (ML) in everyday settings, it is critical to understand the behavior and underlying decision-making of ML models so as to increase people's trust in and acceptance of ML models. To help mitigate the spread of misinformation on social media platforms, explanations have been proposed to help end users understand the fact-checked fake news articles by ML algorithms, and thus change their perception of and belief in fake news articles.

### 2.2 Counterfactual Explanations

According to [7], counterfactual thoughts play an important role in cognition, emotion, and social judgments. People often create counterfactuals about how things could have been changed in the past. They usually imagine a better outcome instead of a worse consequence based on different conditions. Counterfactual reasoning is also considered to be "human-friendly" because its contrastive and selective nature closely resembles human reasoning when people seek to understand the causal structure of events [9, 17, 26]. Counterfactuals are also backed up by psychology: [7] argues that counterfactuals when used expositorily are more persuasive and entertaining. In interpretable ML, to explain the decision made in a given instance, a counterfactual explanation is another instance that is constructed to show the minimal changes that result in a different model decision [27, 42]. Based on the above knowledge, we seek a novel XAI method by which to generate comprehensible and convincing counterfactual explanations for fake news.

### 2.3 Familiarity

People believe in repeated information more than novel information, regardless of its truth or falsity [16, 39]. Such a repetition-induced truth effect (or illusory truth effect) has been explained by people's reliance on familiarity or fluency as a cue for truth judgment [3, 12].

Whereas published studies focus on investigating how repetition affects people's belief in misinformation in experimental [33, 36] and real-world [14] settings, little work has been done to understand how such familiarity impacts the debunking of misinformation. Given the current proliferation of misinformation on social media [41], it is critical to present a corrective message conveying that the presented message is false [10, 11].

In this work, we focus on counterfactual explanations. A counterfactual statement typically simulates two possibilities: the conjecture and its presupposed facts (for a review, see [8]). Thus, the effectiveness of a counterfactual replies on the activation of both possibilities. A recent eye-tracking experiment showed that given a counterfactual, participants looked at the conjecture first, and later the presupposed facts [30]. Moreover, the participants looked primarily at the presupposed facts regardless of whether the counterfactual was affirmative or negative. For a counterfactual explanation of a piece of misinformation, the presupposed fact corresponds to the falsity, and the conjecture reveals key evidence pinpointing the falsity. Depending on the familiarity level of misinformation, the activation of presupposed facts may vary among individuals, consequently impacting the effectiveness of counterfactual explanations. Thus, we examine people's prior knowledge about misinformation and evaluate how this impacts the effect of counterfactual explanations. Considering that the presupposed facts are *specific* for negative counterfactual explanations, but not for affirmative counterfactual explanations, in this work we also compare different counterfactual types.

## 3 METHODOLOGY

When generating a counterfactual explanation, the task is to generate a sentence explaining the counterfactual, given a false claim and a piece of corresponding verified evidence that refutes the false claim. Figure 1 illustrates the three steps of our counterfactual explanation generation framework. First, we adopt a modified Questgen AI[2] as the question and answer generator which we use to generate several keyword-based question-answer pairs. Each question corresponds to a keyword extracted from the false claim. Then, the questions are answered with the relevant human-annotated evidence. Second, we use an entailment checker to pick the best answer that is the most contradictory to the false claim. Finally, we convert the best answer to a complete sentence with a QA-to-claim model and apply our counterfactual templates. Below we describe each step in detail.

### 3.1 Asking Good Questions: Question-Answering Generator

**Overview** The goal of the question-answer generator is to retrieve the possible misleading description from the false claims and answer the questions with the verified evidence to support the refutation of the counterfactual explanations.

Although previous retrieval-based question generators usually pick noun phrases or entities as the question topic candidates, we find that misleading information exists not only in noun phrases or entities. That is, in the real world, fake claims contain various forms of misleading information. We manually analyze the refuted claims in FEVER [38], and find that most misleading claims contain not only incorrect entities and name phrases but also incorrect name entities, adjectives, numbers, dates, or even adverbs. We thus extend the question topic candidates with additional parts of speech to help the question generator better focus on possible misinformation in the false claims.

We construct the question-answer generator based on the open-source Questgen AI template [15], a T5-based QA-generator fine-tuned on QQP,[3] SQuAD [34], and MSMarco [28]. The original setting extracts only name entities, name phrases, and pronouns as candidate keywords. Based on previous arguments, we define the POS settings to pick candidate keywords from nouns, pronouns, adjectives, numbers, determiners, and adverbs. With this extension, we assist the model to focus on a greater variety of keywords and further generate different kinds of keyword-based questions. The generation of questions based on the specified parts of speech in false claims helps to identify statements in the sentences which could be incorrect.

---
[2]https://github.com/ramsrigouthamg/Questgen.ai
[3]https://www.kaggle.com/c/quora-question-pairs/data

### 3.1.1 Question Generation.
The proposed question generation module consists of a standard encoder-decoder T5 model. Question generation involves first extracting at most ten keywords from the false claim using `pke` [4], a python-based implementation of state-of-the-art keyphrase extraction models. We restrict the parts of speech of the candidate keywords to PROPN, NOUN, ADJ, DET, NUM, and ADV.[4] For each keyword $K_i$, the question generator generates questions $Q_i = \mathcal{G}(Q, K_i)$ by treating $K_i$ as an answer and the false claim as the context. As there may be multiple keywords extracted from a single claim, we map all keywords to the corresponding claim and then input all keyword-claim pairs. The encoder represents the claim and answer in a single packed sequence of tokens, using special tokens `context:` and `answer:` to separate the answer from the claim and `</s>` at the end of a input sequence. The complete input is thus `context: $C_i$ answer: $K_i$ </s>`. The encoder then represents the input sequence as a single fixed-length continuous vector. This vector representation is passed to the decoder to generate the questions. We collect all questions and filter out duplicate questions.[5]

### 3.1.2 Answer Generation.
The answer generator then takes the human-annotated verified evidence $E$ from the dataset and the generated questions $Q_i$ as inputs to generate the answers $A_i$ to each question set. The model structure is the same as for question generation, which consists of a standard T5 model. We input our generated questions and the evidence in a single packed sequence of tokens. We use special tokens `question:` and `<s> context:` to separate the question from the evidence and `</s>` at the end of a input sequence. The resulting complete input sequence is thus `question: $Q_i$ <s> context: $E$ </s>`. The encoder-decoder structure is the same as that for the question generator; we use greedy decoding to convert IDs to the output answers with a maximum length of 256. Finally, we strip and capitalize the predicted sentence.

## 3.2 Finding Contradictions: Entailment Checker

In this step, we select the best answer to refute the input claim, which we define as the most contradictory answer to the claim. We implement an entailment checker based on the RoBERTa-large Model [24], and fine-tune the model on MultiNLI.[6] We input each false claim $C_i$ with the generated answers $A_i = \{a_0^i, a_1^i, ..., a_k^i\}$ from the previous step. The false claim and answer are separated by special tokens `</s></s>` to a single input sequence. As there are $k$ answer pairs with each claim $C_i$, we collect a set of $k$ input sequences and feed these into the encoder. The entailment checker then outputs the predicted logits $L_i = \{l_0^i, l_1^i, ..., l_k^i\}$ for each claim-answer pair. Every $l_k^i$ is a triple of logits corresponding to three labels (Contradiction, Neutral, and Entailment): a higher logit value represents higher confidence with respect to that label. We select the answer with the highest Contradiction logit as the best answer. We expect the final answer to have the most contradictory semantic meaning compared to the claim, that is, the best answer that refutes the claim.

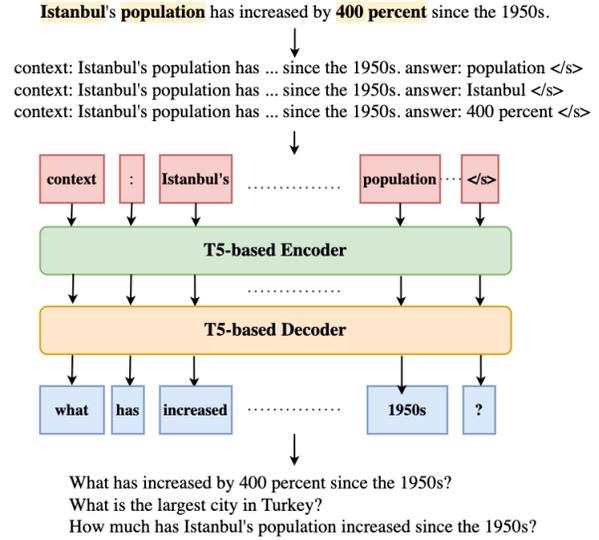

Figure 2: Architecture of proposed question generator. Keywords extracted from the claim are highlighted in yellow. For each extracted keyword, we generate a single input sequence and feed it into the encoder one-by-one. After the decoder generates the questions, we capitalize the sentences as the final output questions.

## 3.3 Reasoning in a Good Way: Explanation Generation

Finally, we generate counterfactual explanations from the generated question-answer pairs and the false claim. Note that the best answers may be not complete sentences but merely short answers. Thus we label the dependency parsing of each answer with spaCy,[7] and tag answers with incomplete sentence structure. If the best answer is labeled as incomplete, we adopt the QA-to-claim model described in [32]. Otherwise, we keep the original best answer with a complete sentence structure. The QA-to-claim model is a BART model [22] fine-tuned on SQuAD [34] and QA2D [13]. The model takes the question-answer pairs $(Q, A)$ as inputs and outputs the declarative sentences. For example, the output of the $(Q_k, A_k)$ pair— $Q_k$: "Where is Stranger Things set?", $A_k$: "Hawkins, Indiana."—is $S_k$: "Stranger Things is set in Hawkins, Indiana." We use special token `[SEP]` to separate the question from the evidence. Other encoder-decoder details are as those described in Section 3.1.2.

---

[4]We use MultipartiteRank as the keyphrase extraction model, which is the implementation of [5]. Clustering arguments settings are alpha = 1, threshold = 0.75, with average clustering.
[5]Implementation details: the embedding size of the encoder and decoder tokens of the question-answer generator is 256, and all inputs are padded to a fixed continuous vector of length 256.
[6]https://huggingface.co/roberta-large-mnli
[7]https://github.com/explosion/spaCy

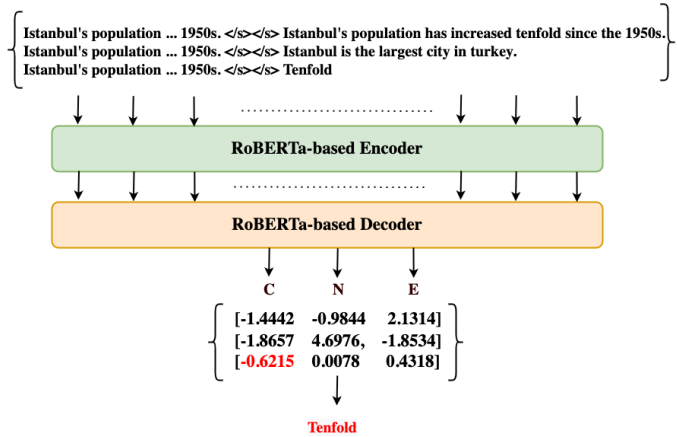

Figure 3: Graphical representation of proposed entailment checker. The decoder outputs the contradiction, neutral, and entailment logits. The highest Contradiction logit is highlighted in red. In this case, the third answer (**Tenfold**) is selected as the best answer.

Then, we follow the counterfactual definition [9] to define our explanation templates. Since counterfactual explanations are created by adding new information rather than deleting misinformation, we link $S_i$ back to false claim $C_i$ to find the smallest change needed to $C_i$ to flip the stand. Inspired by [29], which clarifies the similarities and differences between negative counterfactuals and affirmative counterfactuals, we define three counterfactual explanation forms. Table 3 shows the results for counterfactual explanations with the following three different forms.

(1) **Counterfactual Explanation: Affirmative** We define the form as "**If we were to say $S_i$ instead of $F_i$, the claim would be correct**." We denote $S_i$ as the declarative sentence and $F_i$ as the smallest change needed to $C_i$ to flip the reader's opinion. We keep this smallest change by removing the same name entities or bi-grams that appear in $S_i$ and $C_i$; this step helps the reader to focus on the core of the false claim.
(2) **Counterfactual Explanation: Negative** In the negative form, we combine the false claim and the correct declarative sentence without any modification to the negative form. We define the form as "**If we were to say not $C_i$ but instead $S_i$, the claim would be correct**." We denote $C_i$ as the false claim sentence and $S_i$ as the declarative sentence.
(3) **Counterfactual Explanation: Mixed** The third form mixes the features from the previous two templates. We define the form as "**If we were to say $NC_i$ and/but say $F_i$, the claim would be correct**." $NC_i$ denotes the negation of the false claim and $F_i$ the smallest change needed to $C_i$ to flip the reader's opinion. To ensure the fluency of the sentence, if $NC_i$ contains negative words e.g., *not*, we structure the sentence as **not…but**. Otherwise, we use *and* to connect $NC_i$ and $F_i$. The false claims are negated through a set of rules based on English grammar.

| Annotated examples | 500 |
| --- | --- |
| Number of input error | 42 (8.4%) |
| Number of system error | 108 (21.6%) |
| Number of overall error | 150 (30%) |

Table 1: Statistics of labeled data

*Results.* Table 1 shows the explanation generation results on a randomly sampled set of 500 sentences. We manually labeled the explanations. An explanation is considered correct when it has the correct grammar and provides a sufficient statement to refute the false claim. About 70% of the explanations of the false claims are labeled valid. However, if we remove the errors from the dataset—e.g., incorrect claim labels, not enough evidence—about 76.4% of the explanations are labeled valid.

| | | |
| --- | --- | --- |
| System error | Answer not correctly picked | 25 (16.7%) |
| | Wrong grammar | 9 (6%) |
| | Wrong answer/question | 74 (43.7%) |
| Dataset error | Wrong claim label | 6 (4%) |
| | Insufficient evidence | 36 (24%) |
| Total error | | 150 |

Table 2: Error analysis of labeled data

Finally, we provide a detailed analysis of the incorrect explanations in Table 2. We divide the 150 incorrect explanations (out of the 500 annotated random samples) into the system error and dataset error categories. *System error* refers to a mistake in any part of the explanation generation pipeline, and *dataset error* refers to annotation error or evidence error from the FEVER dataset.

To gain a richer perspective on the system error, we annotated all incorrect explanations of system error with five sub-categories. We find that 43.7% of the incorrect cases come from the question and answer generation, which means that not all questions and answers generated from this step refute the false claim. Among these 43.7% incorrect explanations, about 2.9% require math, and 5.4% require commonsense to complete the explanations. For example, some false claims include incorrect time periods or incorrect statements of death; in these the evidence indirectly indicates the mistake. In this case, the answering model fails to answer correctly due to a lack of math and commonsense ability.

**Answer not correctly picked** accounts for 16.7%: for this error type, the entailment checker fails to pick the answer that refutes the claim but instead picks an irrelevant answer. We further find that not all correct answers contain the most contradictory semantic meaning compared to the false claim. That is, some misinformation in the claim has neutral semantic meaning compared to the correct information, e.g., chess club and game company. As a result, our assumption—that the best answer must have the most contradictory semantic meaning—does not hold in this case.

Our current framework generates a complete counterfactual explanation in a clear, fluent way with 70% overall correctness. However, there are still some limitations in our framework. First, some evidence provides indirect or insufficient information to refute the claim.

| | |
|---|---|
| **False claim** | Black Panther is exclusively a comic book. |
| **Declarative sentence** | Black panther is an upcoming American superhero film based on the marvel comics character of the same name. |
| CF-A | If we were to say 'Black panther is an upcoming American superhero film based on the marvel comics character of the same name' instead of 'exclusively a comic book', the claim would be correct. |
| CF-N | If we were to say not 'Black Panther is exclusively a comic book' but instead 'Black panther is an upcoming American superhero film based on the marvel comics character of the same name', the claim would be correct. |
| CF-M | If we were to say 'Black Panther is NOT exclusively a comic book but an upcoming American superhero film based on the marvel comics character of the same name', the claim would be correct. |
| EXT | Black Panther is set to be released in the United States on February 16, 2018, in IMAX. |
| ABS | Black Panther is an upcoming American superhero film based on the Marvel Comics character of the same name. |
| **False claim** | LGBT is not an acronym containing the word lesbian. |
| **Declarative sentence** | LGBT is an initialism that stands for lesbian, gay, bisexual, and transgender. |
| CF-A | If we were to say 'LGBT is an initialism that stands for lesbian, gay, bisexual, and transgender' instead of 'not an acronym containing the word lesbian', the claim would be correct. |
| CF-N | If we were to say not 'LGBT is not an acronym containing the word lesbian' but instead 'LGBT is an initialism that stands for lesbian, gay, bisexual, and transgender', the claim would be correct. |
| CF-M | If we were to say 'LGBT is an acronym containing the word lesbian and an initialism that stands for lesbian, gay, bisexual, and transgender', the claim would be correct. |
| EXT | The initialism, LGBT, is intended to emphasize a diversity of sexuality and gender identity-based cultures. |
| ABS | It is an adaptation of the initialism LGB, which was used to replace the term gay in reference to the LGBT community. |

Table 3: System output on two randomly sampled sentences given the false claim and declarative sentence. CF-A, CF-N, CF-M, EXT, and ABS denote a counterfactual explanation (Affirmative), a counterfactual explanation (Negative), a counterfactual explanation (Mixed), a extractive-based summarization explanation, and a abstractive-based summarization explanation respectively.

Second, we believe that some situations require that the answer generation step acquire extra knowledge, commonsense reasoning, numerical analysis, or even extract answers considering discourse relation.

## 4 HUMAN EVALUATION

We conducted two online surveys on Amazon Mechanical Turk (MTurk) to examine the effectiveness of counterfactual explanations in helping people understand why a piece of information is fake. In *Survey-1*, we proposed three counterfactual explanations: affirmative (**CF-A**), negative (**CF-N**), and mixed (**CF-M**), and compared their explainability in terms of why a piece of news is fake (**RQ1**). We also measured participants' prior knowledge (familiar vs. unfamiliar) with fake claims and evaluated its influence on the counterfactual explanations (**RQ3**). In *Survey-2*, we compared the best counterfactual explanation from *Survey-1* with two state-of-the-art summary-based explanations: extractive (**EXT**) and abstractive (**ABS**, **RQ2**).[8] The effect of familiarity was also evaluated (**RQ3**).

We restricted both surveys to participants who (1) were at least 18 years old; (2) had completed more than 100 human intelligence tasks (HITs) with a HIT approval rate of at least 95%; and (3) were located in the United States. We also obtained approval from the institutional review board (IRB) office at the authors' institution before commencing the study. [9]

### 4.1 Results

Both surveys took about 20 min to complete. In *Survey-1* and *Survey-2*, 33 and 54 participants failed the attention check, respectively. We did not prevent participants from selecting the same explanation for the best and the worst options, since it was intuitive as long as the participants read the instructions. Nevertheless, for 217 (*Survey-1*) and 594 (*Survey-2*) responses, participants selected the same explanation as the best and the worst options. We excluded the above-mentioned responses from the following data analysis. Each participant's responses to the first fake claim and associated explanations were treated as practice trials, which were also excluded from data analysis.

*Survey-1.* The *Survey-1* results are shown in Table 4.[10] Among the three counterfactual explanations, **CF-A** was the most selected best explanation ($\chi^2_{(2)} = 21.95, p < 0.001$). The effect of familiarity showed a significant impact on the best explanation selection ($\chi^2_{(2)} = 11.16, p = 0.004$). However, results of the worst explanation showed no significant differences ($\chi^2_{(2)} = 3.20, p = 0.202$).

---
[8]The detailed information of baseline models could be found in Appendix A.2.

[9]We describe the materials, and procedures of the two survey in Appendix section B.
[10]To examine the effect of the three counterfactual explanations, we only selected claims and explanations from the samples without system error.

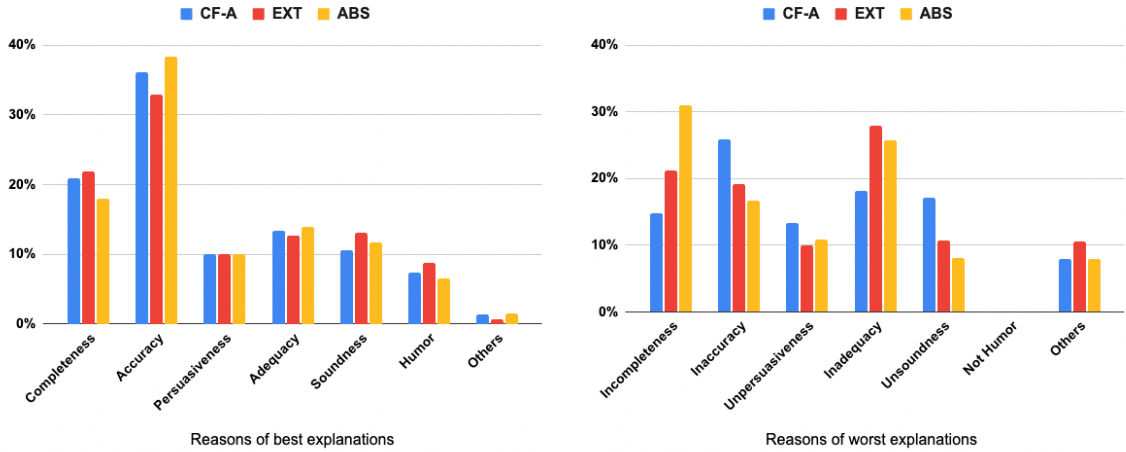

**Figure 4: Aspect analysis results of *Survey-2*.**

Interestingly, participants' familiarity with fake claims also impacted their perceived accuracy evaluation. While participants who were familiar with fake claims judged the fake claims as real more (416/480, 86.7%), the opposite pattern was evident for participants who were unfamiliar with the fake claims (242/346, 69.9%).

| Model | Best | | | | Overall (988) | Worst | | | | Overall (988) |
|---|---|---|---|---|---|---|---|---|---|---|
| | Familiar (581) | | Unfamiliar (407) | | | Familiar (581) | | Unfamiliar (407) | | |
| | PR (513) | PF (68) | PR (120) | PF (287) | | PR (513) | PF (68) | PR (120) | PF (287) | |
| CF-A | **0.41** | 0.29 | **0.42** | **0.40** | **0.40** | 0.31 | **0.40** | 0.31 | 0.27 | 0.31 |
| CF-N | 0.32 | **0.35** | 0.28 | 0.33 | 0.32 | 0.34 | 0.24 | **0.37** | 0.33 | 0.34 |
| CF-M | 0.27 | **0.35** | 0.31 | 0.26 | 0.28 | **0.35** | 0.37 | 0.33 | **0.35** | 0.34 |

**Table 4: Result of *Survey-1*.** Proportion of each explanation being selected as the best or the worst explanation. The best performance in each column is in boldface. Familiar and Unfamiliar show responses from participants who recognized and did not recognize the claims, respectively. Participants also evaluated the perceived accuracy of fake claims before the explanation evaluation. *PR* indicates the number of responses *P*erceived as *R*eal and *PF* indicates the number of responses *P*erceived as *F*ake. The number in parentheses under each tag is the size of each group.

*Survey-2*. Based on the results of *Survey-1*, we compared **CF-A** with the **EXT** and **ABS** explanations in *Survey-2*. To mimic the effectiveness of the fact-checked explanations in a real-world scenario, we did not filter out samples with system errors. Table 5 shows the results. The best performance was evident for the counterfactual explanation (**CF-A**) in both the best selection ($\chi^2_{(2)} = 51.84, p < 0.001$) and the worst selection ($\chi^2_{(2)} = 6.46, p = 0.039$). In agreement with the results of *Survey-1*, the best option results were also dependent on participants' prior knowledge of the fake claims ($\chi^2_{(2)} = 15.39, p < 0.001$). To understand the effect of system error, we conducted the same analyses by excluding system errors (30%) and obtained the same results (see Table 6).

Consistent with the results of *Survey-1*, participants' familiarity with fake claims impacted their perceived accuracy evaluation regardless of system errors.

### 4.2 Discussion

**RQ1: How to generate a good counterfactual explanation for the providing fake claim?**

We have shown our proposed method in the Methodology section, which consists of three steps of to generate counterfactual explanations: *Step 1*. QA generation; *Step 2*. Entailment Checking; and *Step 3*. Explanation Generation. With such method, we achieved an overall correctness of 70%. As shown in Tables 5, 6, and 7, **CF-A** showed an overall better performance compared to the other baseline explanations.

**RQ2: Are different types of counterfactual explanations (e.g., affirmative, negative, and mixed) varying in best explaining why a piece of news is fake?**

Table 4 shows the result of *Survey-1*. **CF-A** was the most frequently selected best explanation among the three forms of counterfactual explanations. We compared the three forms of explanations. A conjecture on the explanation for such gap is because **CF-A** starts

| Model | Best | | | | Overall (965) | Worst | | | | Overall (965) |
|---|---|---|---|---|---|---|---|---|---|---|
| | Familiar (480) | | Unfamiliar (485) | | | Familiar (480) | | Unfamiliar (485) | | |
| | PR (416) | PF (64) | PR (136) | PF (349) | | PR (416) | PF (64) | PR (136) | PF (349) | |
| CF-A | **0.42** | **0.38** | **0.41** | **0.49** | **0.44** | 0.29 | 0.38 | **0.40** | 0.27 | 0.25 |
| EXT | 0.27 | 0.28 | 0.30 | 0.27 | 0.27 | **0.38** | 0.30 | 0.27 | 0.29 | **0.40** |
| ABS | 0.31 | 0.34 | 0.29 | 0.24 | 0.28 | 0.33 | 0.33 | 0.32 | **0.44** | 0.35 |

**Table 5: Result of *Survey-2*.** CF-A, EXT, ABS denote counterfactual explanation (Affirmative), extractive-summarized explanation and abstractive-summarized explanation respectively.

| Model | Best | | | | | Worst | | | | |
|---|---|---|---|---|---|---|---|---|---|---|
| | Familiar (337) | | Unfamiliar (346) | | Overall (683) | Familiar (337) | | Unfamiliar (346) | | Overall (683) |
| | PR (296) | PF (41) | PR (104) | PF (242) | | PR (296) | PF (41) | PR (104) | PF (242) | |
| CF-A* | 0.39 | 0.41 | 0.41 | 0.52 | 0.44 | 0.31 | 0.29 | 0.40 | 0.24 | 0.26 |
| EXT | 0.28 | 0.29 | 0.33 | 0.26 | 0.28 | 0.37 | 0.37 | **0.25** | 0.29 | 0.40 |
| ABS | 0.33 | 0.29 | 0.26 | **0.21** | 0.28 | 0.32 | 0.34 | 0.35 | 0.47 | 0.35 |

**Table 6:** *Survey-2* **results of without any counterfactual generation system errors.** We manually annotate all random samples and filtered out 30 % of the samples. The result shows an insignificant difference with Table 5.

| Model | average ranking | | | | | average ranking* | | | | |
|---|---|---|---|---|---|---|---|---|---|---|
| | Familiar (480) | | Unfamiliar (485) | | Overall (965) | Familiar (480) | | Unfamiliar (485) | | Overall (683) |
| | PR (416) | PF (64) | PR (136) | PF (349) | | PR (416) | PF (64) | PR (136) | PF (349) | |
| CF-A | **1.86** | 2.0 | 1.99 | **1.78** | **1.86** | **1.92** | **1.87** | 1.99 | **1.72** | **1.86** |
| EXT | 2.11 | 2.01 | **1.97** | 2.03 | 2.05 | 2.09 | 2.07 | **1.92** | 2.02 | 2.03 |
| ABS | 2.02 | **1.98** | 2.04 | 2.20 | 2.08 | 1.98 | 2.05 | 2.09 | 2.25 | 2.10 |

**Table 7: Ranking-based result of** *Survey-2*. The average ranking* calculates the average ranking without any counterfactual generation system errors.

with new evidence while both **CF-N** and **CF-M** start with negation of the fake claims.

**RQ3: How do counterfactual explanation best explain why a piece of news is fake compared to other state-of-the-art explanations?** In *Survey 2*, we asked the participants to select which aspect they think the selected explanation is the best and answer why they selected the explanation as the best using an open-ended questions. For the selected worst explanation, we asked the same questions. Figure 4 shows the results of the aspects selection. From the results, we can intuitively argue that accuracy is the most important aspect for a great explanation. In contrast, incompleteness is the most important aspect for a bad explanation.

Moreover, participants' responses to the open-ended questions corroborate their selection of the best and the worst explanations. We found that most participants indicated that counterfactual explanations can accurately explain why a piece of news is fake, including *"The explanation covers everything about the claim."* and *"This one seemed to make the most sense out of all of them."* Meanwhile, participants also provided insights as to why selecting **EXT** and **ABS** explanations as the worst. For the **EXT** explanation, participants replied *"It doesn't answer the question or refute the claim."* or *"It does not address the issue."* For the **ABS** explanation, participants indicated that *"this explanation makes no sense"* and *"This doesn't answer the claim at all."* and *"This says nothing about the material."*

One possible explanation for the above findings is that the experiment is run on FEVER, all the evidence documents of which are from Wikipedia. The summarized-based model may focus on other statements that the model thinks are more important to represent the whole evidence document. Such results also align with previous findings [2]. Nevertheless, our proposed framework would extract multiple questions from the evidence document, thus our generated explanations may have the least possibility to miss the arguments of the fake news claims.

**RQ4: Does individuals' familiarity (familiar vs. unfamiliar) with misinformation impact the effectiveness of counterfactual explanations?** We asked subjects whether they had seen or heard the claims before reading the explanations. In *Survey-2*, about similar numbers of participants were familiar or unfamiliar with the fake claims. As shown in Table 6, **CF-A** achieve the best performance in all the conditions. Fifty-two percent of the participants who were unfamiliar with the fake claims selected **CF-A** as the best explanation. If we examine the effect using all samples, the proportion of **CF-A** was 49%. The results also show the limited impact of familiarity on the effectiveness of **CF-A** explaining why a piece of news is fake. The **CF-A** were selected as the best explanation by both familiar ($\chi^2_{(2)} = 15.39, p < 0.001$) and unfamiliar ($\chi^2_{(2)} = 41.2, p < 0.001$) groups.

## 5 LIMITATIONS

*Dataset.* In this paper, we only demonstrated the result on the FEVER dataset. Even though FEVER is a widely used dataset for fake news-related research, FEVER was constructed with the articles from Wikipedia, which might have different results while using news articles.

*Source of human subjects.* We only recruited MTurk workers, who tend to be young and well educated [19]. Although the MTurk workers are demographically diverse, they are not nationally representative samples [6]. Future work could consider a more comprehensive recruitment to generalize our findings to other samples (e.g., recruiting representative samples or participants from other platforms such as Prolific [31]).

*Model.* Since our method is a pipeline framework, each process influence the final generated explanation. For example, if the model cannot find the correct question-answer pair in step 1, QA Generation, the ultimate result will be inherently wrong. Thus, in our approach, each process is essential. If any process did not work well, the generated counterfactual explanation would be incorrect.

## 6 CONCLUSION

This work present a method on generating counterfactual explanations. We showed that counterfactual explanations outperform other summarized-based explanations considering people's initial stance toward the fake claims. This work also contributes to understanding how different form of counterfactual explanations may affect people's sense toward the counterfactual explanations.

For future work, other approaches of generating counterfactual explanations could be investigated. Especially improving the ability of acquiring extra knowledge, commonsense reasoning or numerical aptitude, as these could increase the performance and the possibility to apply in the real world.

## ACKNOWLEDGMENTS

This research are supported by Ministry of Science and Technology, Taiwan, under Grant no. 110-2221-E-001-001- and 110-2634-F-002-051-. Further, this work is supported by Academia Sinica under AS IIS 3006-37C4218. The works of Aiping Xiong were in part supported by NSF award 1915801.

## A EXPERIMENT

### A.1 Dataset

We conducted experiments on the FEVER [38] training dataset, which contains annotated claim-evidence pairs. We generated counterfactual explanation sentences using our framework.

**FEVER** consists of more than 185K claim-evidence pairs. The claims are generated by altering sentences extracted from Wikipedia;[11] evidence is constituted by manually verified sentences from the corresponding Wikipedia pages. Each pair is classified as SUPPORTED, REFUTED, or NOTENOUGHINFO. In the following experiments, we randomly sampled 500 sentence pairs from the 29,351 refuted claim-evidence pairs labeled REFUTED.[12]

### A.2 Baselines

Two studies generate fact-checked explanations. Both methods are summarization-based methods. [2] proposes an extractive-based summarization model for extracting explanations from evidence documents, and [21] propose a system that generates explanations by use of an abstractive-based summarization model. As the authors did not release the model source code, we attempted to reproduce the models according to the information mentioned in their studies.

**Extractive Summarization** For the extractive summarization model, we used the uncased pre-trained DistillBert. The model was pre-trained on the CNN/Daily Mail dataset for 3 epochs.

**Abstractive Summarization** For the abstractive summarization model, the authors adopted [23]. The goal was to generate a fact-checked explanation for health-related fake news. Thus, the authors pre-trained their summarization model on a health-related dataset. In our task, we focus on the FEVER dataset; FEVER contains a variety of topics not limited to public health. Thus, in this abstractive explanation setting, our work is based on an off-the-shelf pre-trained abstractive summarization model.[13] This is a RoBERTa Transformer pre-trained on the CNN/Daily Mail dataset.

We compare counterfactual explanations with extractive and abstractive explanations. We selected these two works because these were the existing state-of-the-art methods for generating fact-checked explanations. We generated both extractive and abstractive explanations by summarizing the evidence of FEVER's claim.

## B HUMAN EVALUATION

### B.1 Participants

We recruited 425 and 625 MTurker workers for *Survey-1* and *Survey-2*, respectively. The demographic information of the participants of both surveys are similar and are shown in Table 8.

### B.2 Materials

We randomly sampled 500 texts from the FEVER dataset to obtain 500 generated counterfactual explanations. We annotated the correctness of these 500 random samples for further analysis. For each sample, we have three different explanations for comparisons on one fake claim. All explanations are limited to one sentence to ensure a similiar amount of information.

[11]https://en.wikipedia.org/
[12]From the total 29,563 pairs, we dropped 212 incomplete claim-evidence pairs that contained no evidence.
[13]https://huggingface.co/patrickvonplaten/longformer2roberta-cnn_dailymail-fp16

| Item | Options | *Survey-1* | *Survey-2* |
|---|---|---|---|
| Gender | Female | 37.8% | 29.7% |
| | Male | 61.9% | 70.1% |
| | Other | 0% | 0.2% |
| | Prefer not to answer | 0.3% | 0% |
| Age | 18–24 | 3.3% | 3.8% |
| | 25–34 | 54.5% | 45.6% |
| | 35–44 | 29.4% | 38.5% |
| | 45–54 | 10% | 7.9% |
| | Over 55 | 2.8% | 4.2% |
| Political leaning | Very Liberal | 16.7% | 15.9% |
| | Liberal | 24.8% | 27.4% |
| | Moderate | 23.9% | 18.6% |
| | Conservative | 19.7% | 22.5% |
| | Very Conservative | 14.9% | 14.8% |
| | Prefer Not To Answer | 0% | 0.8% |
| Ethnicity | American India or Alaska Native | 0.6% | 0.4% |
| | African or African American | 10.3% | 14.8% |
| | Native Hawaiian or Pacific Islander | 0% | 0.7% |
| | White | 73.2% | 67.3% |
| | Hispanic or Latino | 3.6% | 3% |
| | Caucasian | 9% | 10.7% |
| | Asian | 1.5% | 1.8% |
| | More than one race | 1.3% | 1.1% |
| | Prefer not to answer | 0.5% | 0.2% |
| Education | Less than high school degree | 0% | 0.4% |
| | High school | 2.8% | 8.1% |
| | Some college but no degree | 3.4% | 5% |
| | Associate degree in college (2-year) | 2.1% | 5% |
| | Bachelor's degree | 66.5% | 50.1% |
| | Master's degree | 23.2% | 29.1% |
| | Doctoral degree | 0.5% | 1.1% |
| | Professional degree (JD,MD) | 1.3% | 1.2% |
| | Prefer not to answer | 0.2% | 0% |

Table 8: Demographic information of participants in *Survey-1* and *Survey-2*.

### B.3 Procedures

After informed consent, *Survey-1* began. Each participant evaluated *five* false claims and the associated explanations. For each claim, participants first indicated whether they had seen or heard it before, and evaluated the perceived accuracy of the claim on a 5-point scale ("1" meaning "very inaccurate" and "5" meaning "very accurate"). We then informed participants that the claim has been classified as fake and asked them to compare the three explanations (i.e., **CF-A**, **CF-N**, and **CF-M**) of why the claim is fake. Specifically, participants first viewed the three explanations and selected the best explanation. Then, we presented the three explanations in the same order and asked participants select the worst explanations. Participants also selected one aspect of the explanation as the main reason for their decisions. They also described why they believed that explanation was the best or the worst using open-ended questions. Half of the participants selected the best explanation first and the other half selected the worst explanation first. The three explanations were presented in a randomized order among participants. To conclude, participants answered several post-session questions, such as their demographic information. Since participants' attention to the claims and explanation were critical to our surveys, we implemented an attention check. Specifically, we informed participants of the attention check and asked them to answer a question with a predefined answer. We paid each participant USD 2.50 for completing the survey.

The procedure of *Survey-2* was the same as *Survey-1* except that participants made the best and worst selections among the best counterfactual explanation from *Survey-1*, the **EXT**, and the **ABS** explanations.

## C ETHICS STATEMENT

We conducted human evaluation with Amazon Mechanical Turk (AMT). We did not ask about privacy information during the annotation process. Participants receive $ 2.5 per HIT through the Turk platform. (The payment includes Amazon Turk handling fee $ 0.01) The payment is reasonable considering it takes average 20 seconds to finish a comparison.